\documentclass{article}

\usepackage[final]{nips_2017}

\usepackage[utf8]{inputenc} 
\usepackage[T1]{fontenc}    
\usepackage{hyperref}       
\usepackage{url}            
\usepackage{booktabs}       
\usepackage{amsfonts}       
\usepackage{nicefrac}       
\usepackage{microtype}      

\usepackage{color}

\usepackage{amsmath,multirow,graphicx}

\newcommand{\eg}{e.\,g.,\ }
\newcommand{\ie}{i.\,e.,\ }
\usepackage{caption}
\usepackage{subcaption}
\usepackage{arydshln}

\title{Weakly Supervised One-Shot Detection with Attention Similarity Networks}

\author{
Gil Keren$^1$, 
Maximilian Schmitt$^1$, 
Thomas Kehrenberg$^1$, 
Bj{\"o}rn Schuller$^{1,2}$ \\ \\
$^1$ ZD.B Chair of Embedded Intelligence for Health Care and Wellbeing, \\ University of Augsburg, Germany \\ 
$^2$ GLAM – Group on Language, Audio \& Music, Imperial College London, UK \\
{\tt cruvadom@gmail.com}
}

\begin{document}

\maketitle

\begin{abstract}
Neural network models that are not conditioned on class identities were shown to facilitate knowledge transfer between classes and to be well-suited for one-shot learning tasks. Following this motivation, we further explore and establish such models and present a novel neural network architecture for the task of weakly supervised one-shot detection. Our model is only conditioned on a single exemplar of an unseen class and a larger target example that may or may not contain an instance of the same class as the exemplar. By pairing a Siamese similarity network with an attention mechanism, we design a model that manages to simultaneously identify and localise instances of classes unseen at training time. In experiments with datasets from the computer vision and audio domains, the proposed method considerably outperforms the baseline methods for the weakly supervised one-shot detection task. 
\end{abstract}

\section{Introduction}

Detection models proposed in the last few years have managed to considerably improve the state-of-the-art performance for object detection tasks. However, the success of these proposed models is due to, among other factors, the use of large-scale datasets that contain a large number of labelled examples for a limited number of classes \citep{deng2009imagenet,lin2014microsoft}. Humans, however, are required in daily life to correctly identify and localise a much larger number of classes. 

Models that perform one-shot learning attempt to overcome the dependency on large amounts of class-specific labelled data by generalising from a single exemplar to other members of its object class. 
Recent works have demonstrated empirical success in one-shot classification using a type of learning models that are not conditioned on class identities at training time \citep{DBLP:conf/nips/VinyalsBLKW16,koch2015siamese}.
Specifically, at every training iteration, the model is given a small number of exemplars from $N$ random classes, and a target example of one of the $N$ classes. The model is trained to emit which of the $N$ exemplars is an instance of the same class as the target example. Class labels are not used during the training procedure, but rather only labels that indicate the similarity of the different exemplars and the target example.
At evaluation time, all inputs are instances of classes unseen during training time. Since the model has no class specific parts, classification output is emitted without any additional training on the unseen classes. 

Training a model in the manner described above encourages the model's performance to be independent of specific classes, and facilitates generalisation from classes seen at training time to unseen classes. Moreover, models trained with no conditioning on specific class identities may induce performance gains also in the standard multi-class classification setting, as knowledge transfer between the different classes is facilitated. The above potentially places models with no conditioning on specific class identities as a key paradigm for machine learning, for both the one-shot and the non-one-shot learning settings. In this work, we intend on further exploring and establishing such models, with the goal of simultaneously classifying and localising classes that were unseen at training time. 

In the standard case of neural network models, generalisation to unseen examples may require a large number of examples to be present in training time. 
Training a model with no conditioning on class identities may induce generalisation to unseen classes, but equivalently, this might require a large number of classes to appear during training. 
Indeed, previous success with such models was observed when the number of distinct classes in the training set was large \citep{DBLP:conf/nips/VinyalsBLKW16,koch2015siamese}. 
Therefore, applying this approach for a one-shot localisation task may require a training corpus that contains bounding box information for a very large number of object classes. 

Bounding box labels are in general harder or more expensive to get, making corpora with bounding box information in most cases smaller than ones that only contain weaker labels. Even in domains where large bounding box-rich corpora do exist, one may still strive to improve model performance by making additional use of the normally larger corpora with no bounding box information.
Therefore, instead of relying on a large enough corpus that contains bounding box information, we attempt to perform one-shot detection using weaker labels alone. For the same reasons, previous work was done on learning (non one-shot) object detection using image-level labels alone \citep{DBLP:conf/cvpr/BilenV16,DBLP:conf/bmvc/TehRW16}. In the absence of bounding boxes at training time, context-based attention models might be good candidates for learning localisation information in a weakly supervised manner. Indeed, such models previously demonstrated the ability to focus on relevant parts of the input element, using labels that do not contain any localisation information \citep{DBLP:conf/icml/XuBKCCSZB15,DBLP:conf/nips/ChorowskiBSCB15,DBLP:journals/corr/BahdanauCB14}. 

In this work, we attempt to further explore and establish models with no conditioning on class identities, and present a novel model for the challenging task of weakly supervised one-shot detection, which is a much less explored task compared to its fully-supervised or non-one-shot counterparts. 
Our model takes a single example of a given class, which we name the \emph{exemplar}, and a larger \emph{target example} that may or may not contain an instance of same class as the exemplar. We adapt the Siamese neural networks framework \citep{DBLP:journals/ijprai/BromleyBBGLMSS93,koch2015siamese} and apply it in a convolutional manner, to compute the similarity between the exemplar and every location of the target example.
Instead of using localisation labels as a supervision signal, \eg bounding boxes, we only use binary labels that indicate the existence of a certain object in the example of interest, and learn the localisation information using an attention mechanism.  
To the best of our knowledge, this is the first time a Siamese similarity network is used in tandem with an attention mechanism to learn similarity between different object parts in a weakly supervised manner. 

We experiment with weakly supervised one-shot detection tasks from the audio and computer vision domains. In the audio domain, we perform spoken term detection \citep{DBLP:conf/asru/HazenSW09,DBLP:conf/icassp/ChenPS15,DBLP:conf/asru/ParadaSR09}, where the task is to identify and localise audio keywords appearing in longer audio utterances. 
In the computer vision domain, we augment the Omniglot dataset \citep{lake2015human} to create a one-shot object detection task by pasting a number of the original characters images onto a larger image. We use the Omniglot dataset as it contains a large number of object classes (1623). We leave experimenting with natural images computer vision datasets, where the image variation is larger, to future work. 
Experiments results show that our method manages to classify and localise instances of unseen classes in both the computer vision and audio domains, and considerably outperforms the baseline methods for the weakly supervised one-shot detection tasks. 
The contribution of our work is then twofold: first, we further explore and establish models with no conditioning on class identities as a leading paradigm for one-shot learning and facilitation of knowledge transfer between classes. Second, we present a novel model for weakly-supervised one-shot detection applicable for different domains, that is able to learn the similarity between object parts in a weakly supervised manner, by pairing a Siamese similarity network with an attention mechanism. 

\section{Related Work} \label{sec:related}
We discuss related work that is not otherwise mentioned. 
Other approaches for one-shot learning that appear in the literature include the work presented in \citet{hariharan2016low}, where for a few-shot recognition task a learner tunes its feature representation on a set of base classes that have many training instances, to better perform on classes with a small number of examples. For the task of one-shot detection, A computer vision model for (fully-supervised) few-shot object detection was presented in \citet{dong2017few}. This model is comprised of a pipeline that includes finding additional training examples in a large unlabelled dataset. 

Weakly supervised object detection was also considered in previous work. \citet{DBLP:conf/eccv/WangRHT14} use a pre-trained convolutional neural network (CNN) to describe image regions and then learn object categories as corresponding visual topics. The model in \citet{DBLP:conf/bmvc/TehRW16} computes an attention score for every location in an image. These scores are then combined to one feature vector describing an image, that is used for classifying the image. Localisation is done using the attention weights. Similarly, the work presented in \citet{DBLP:conf/cvpr/BilenV16} starts from a CNN pre-trained for image classification on a large dataset, then computes scores for each class at each location. These scores are combined into a single image level score, and the network is optimised for the classification task. Detection is again performed according to the class location scores. However, all of the above approaches depend on a predefined set of of classes, and are not suited for one-shot detection. In the audio domain, a related model for query-by-example spoken term detection was proposed concurrent to our work \citep{ao2017query}. This model uses an long short-term memory (LSTM) network for scoring the existence of a keyword in different utterance locations. These scores are combined for a single score for the utterance, that is used for training on a binary classification task. 

\begin{figure}
\centering
\begin{subfigure}{.5\textwidth}
  \centering
  \includegraphics[width=.95\linewidth]{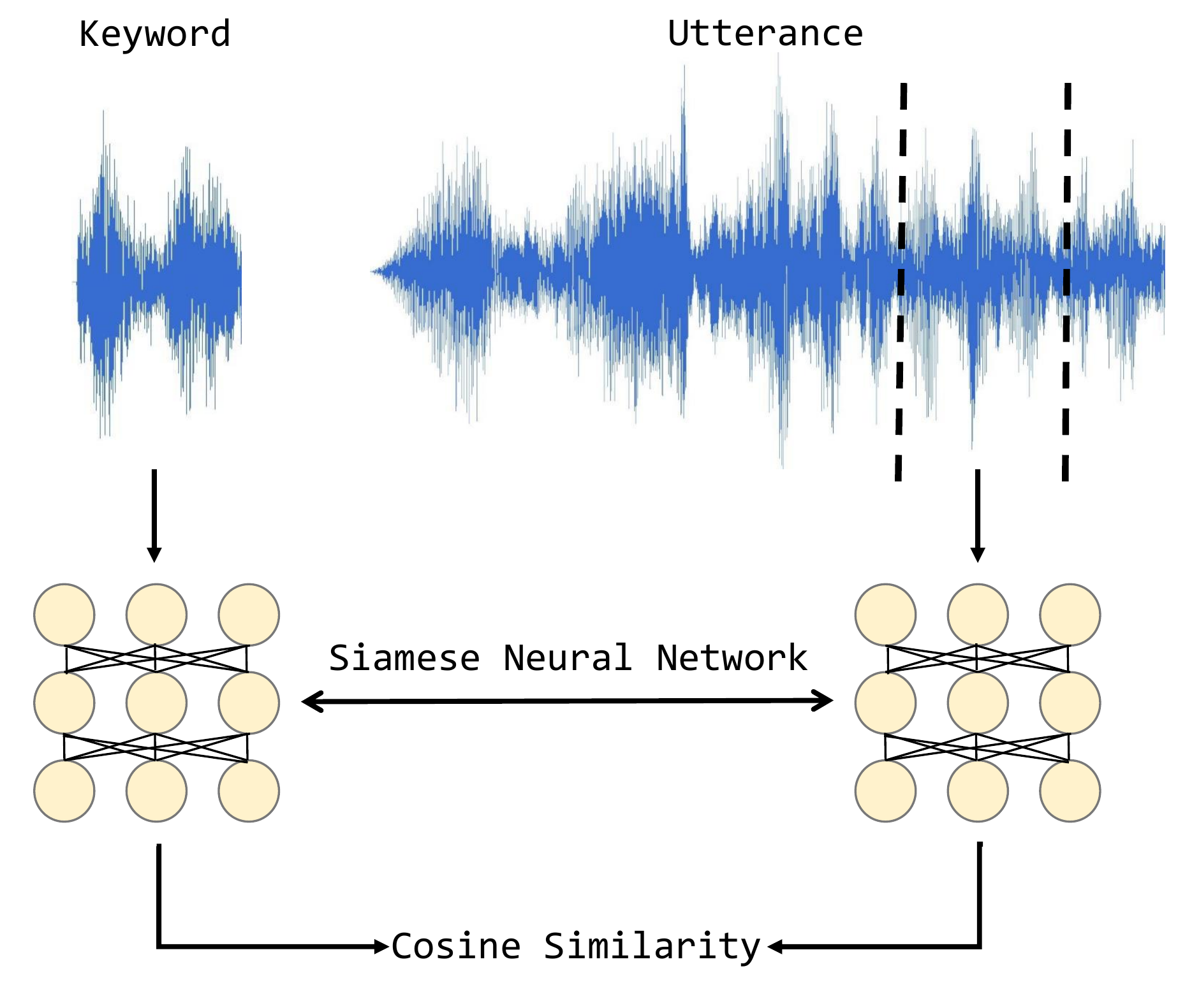}
  \caption{}
  \label{fig:fig1}
\end{subfigure}%
\begin{subfigure}{.5\textwidth}
  \centering
  \includegraphics[width=.95\linewidth]{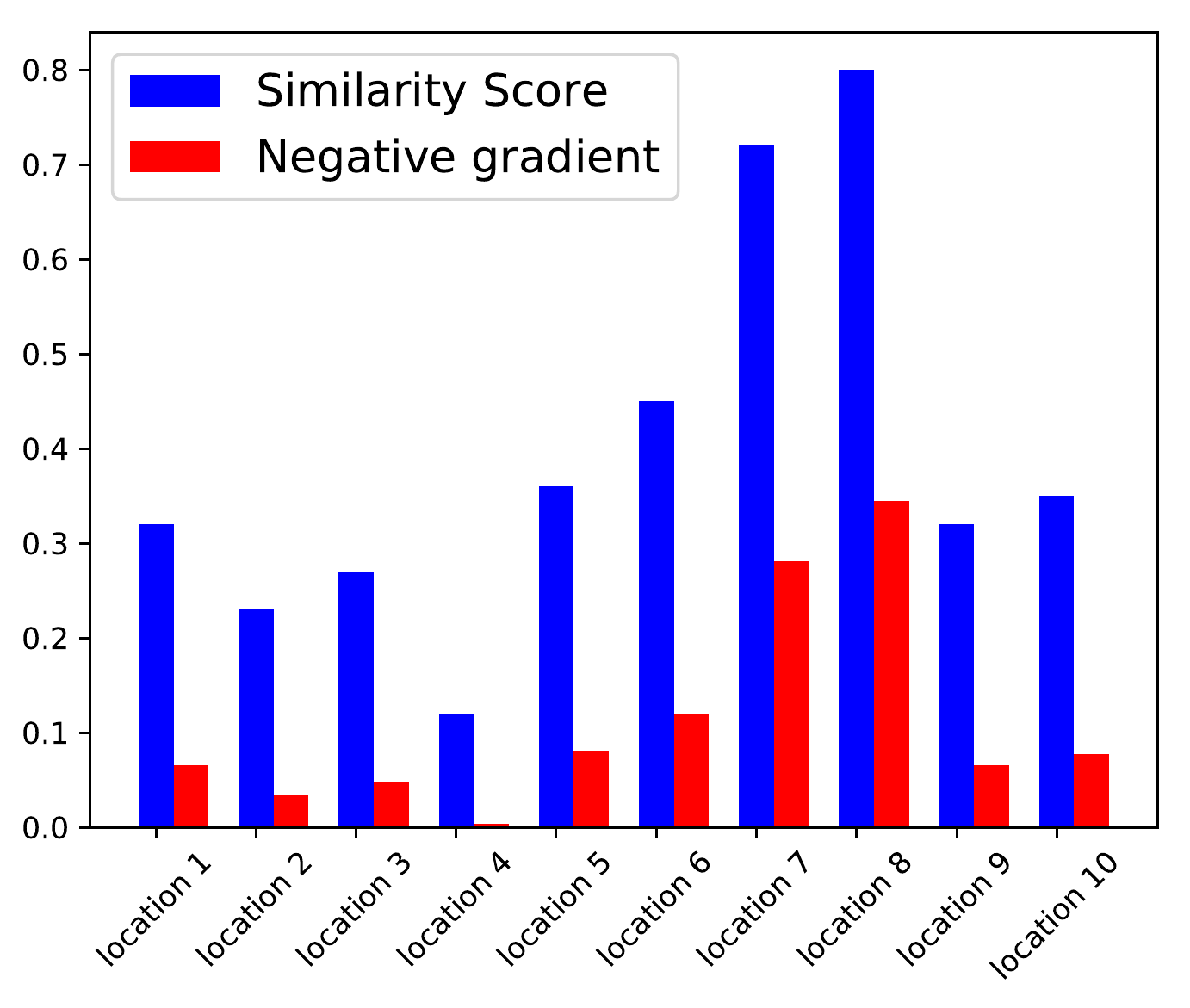}
  \caption{}
  \label{fig:fig2}
\end{subfigure}
\caption{(a) The computation of a similarity score between an exemplar and a target location, for the case of detecting audio keywords in longer utterances. Both the exemplar (keyword) and the target location (part of the utterance) are mapped into a representation space using a Siamese neural network, and a cosine similarity between the representations is calculated. (b) he reason for the self-reinforcing loop: example relation between similarity scores $s_l$ and the negative gradients $-\frac{\partial \ell}{\partial s_l}$, in a positive exemplar-target pair with ten different locations, computed according to Eq. \ref{eq:grad} for an example vector of similarity scores. The negative gradient is increasing with the similarity scores. As a result, locations with high similarity scores will get even higher similarity scores during training, compared to other locations.}
\end{figure}

\section{Method} \label{sec:method}
Our model takes an \emph{exemplar} $x$ -- a single instance of a class of interest and a \emph{target example} $B$, which may or may not contain an instance of the same class as the exemplar. The target example is normally spatially larger than the exemplar, so that the exemplar could be compared to different locations in the target example. We name different locations in the target image \emph{target locations}. The output of the model is a \emph{similarity map} $s(x, B)$ over target locations, such that $s_l(x, B)$ is a measure of the similarity of location $l$ in the target example to the exemplar. For simplicity of notation, we write $s_l$ instead of $s_l(x, B)$ when the context is clear. 

\subsection{Similarity Scores}
To obtain a meaningful similarity measure between the exemplar and target locations, we do not require the raw representations to be within a small distance from one another. Instead, we would like to map both the exemplar and target locations to a latent representation space, and optimise for similarity or dissimilarity in the representation space. Therefore, the computation of $s$ is modelled using the cosine similarity between outputs of a Siamese neural network \citep{DBLP:journals/ijprai/BromleyBBGLMSS93,koch2015siamese}: 
\begin{equation} \label{eq:sim}
s_l = \frac{f_\theta(x) \cdot f_\theta(B_l)}{\| f_\theta(x) \| \| f_\theta(B_l) \|},
\end{equation}
where $B_l$ is location $l$ in the target example $B$, and $f_\theta$ is a neural network parameterised by $\theta$, that is embedding both the exemplar and target locations into the latent embedded space. The similarity score $s_l$ is in the unit interval. See Figure \ref{fig:fig1} for a visualisation of the similarity score computation for one target location. The above computation can be seen as (and was implemented as) a convolutional application of the Siamese network \citep{DBLP:conf/eccv/BertinettoVHVT16} across all locations of the target example, and application of the cosine similarity to the resulting map. In case $f_\theta$ is a CNN, computation can be reduced by applying $f_\theta$ once on the whole target example and extracting the relevant parts of the feature maps that correspond to each location, as was done in \citet{DBLP:conf/iccv/Girshick15}.

\subsection{Weakly Supervised Detection}
We denote every \emph{Siamese pair} $(x, B_l)$ as either positive, if the target location $B_l$ contains an instance of the same class as the exemplar, or negative, otherwise. Similarly, we denote every \emph{exemplar-target} pair $(x, B)$ as either positive, if an instance of the same class as $x$ appears in some location in $B$, or negative, otherwise. The goal of training is to increase similarity scores $s_l(x, B_l)$ for positive Siamese pairs and decrease it for negative Siamese pairs. Recall, that we do not have access to labels containing any localisation information, but rather only binary labels $y_{x, B} \in \{0, 1\}$ that indicate whether the exemplar-target pair $(x, B)$ is positive or negative. 

In the absence of localisation labels, an additional credit assignment problem arises during training -- namely, which Siamese pairs should be assigned a greater similarity score and which should be assigned a smaller similarity score. For negative $(x, B)$ pairs, the answer is simple -- similarity score $s_l$ should be reduced to zero for all locations $l$ (all Siamese pairs are negative). However, for positive $(x, B)$ pairs, the similarity score should be increased for locations that contain the instance of the same class as the exemplar (positive Siamese pairs), and decreased for locations that do not contain such instance (negative Siamese pairs), without any labels that contain location specific information.

We attempt to overcome this issue by making locations in a positive exemplar-target pair compete for a high similarity score, as explained below. Related methods for weakly supervised detection were introduced in \citet{DBLP:conf/cvpr/BilenV16} and \citet{DBLP:conf/bmvc/TehRW16}, but these are not suited for one-shot learning (see Section \ref{sec:oneshot}). We compute attention weights by applying softmax normalisation to the similarity map
\begin{equation} \label{eq:softmax}
w_l = \frac{\exp(s_l / T)}{\sum \limits _{l'} \exp(s_{l'} / T)},
\end{equation}
where $T$ is the softmax temperature. The attention weights are then used to compute a single similarity score for an exemplar-target pair
\begin{equation} \label{eq:yhat}
\hat{y}_{x, B} = \sum \limits _l w_l s_l,
\end{equation}
which is again in the unit interval, and the loss for a single pair $(x, B)$ is
\[ \ell (x, B) = (\hat{y}_{x, B} - y_{x, B})^2. \]
Using the gradient of the softmax function, we can compute
\begin{equation*}
\begin{split}
\frac{\partial \hat{y}}{\partial s_l} & = w_l + \frac{s_l w_l (1-w_l)}{T} - \sum \limits _{l' \neq l} \frac{ s_{l'} w_l w_{l'}}{T} \\
& = w_l (1 + \frac{s_l}{T} - \frac{1}{T} \sum \limits _{l'} w_l s_l ) = w_l (1 + \frac{s_l - \hat{y}_{x,B}}{T}),
\end{split}
\end{equation*}

and the gradient of the above loss function with respect to similarity scores is
\begin{equation} \label{eq:grad}
\frac{\partial \ell}{\partial s_l}(x, B) = 2(\hat{y}_{x,B} - y_{x,B}) w_l (1 + \frac{s_l - \hat{y}_{x,B}}{T}).
\end{equation}
For positive exemplar-target pairs ($y_{x,B} = 1$), the gradient in Eq. \ref{eq:grad} is monotonic decreasing in $s_l$ and $w_l$. Therefore, as the step size is proportional to the negative gradient, propagating this gradient back to network parameters during training will cause similarity scores for Siamese pairs that already have higher similarity scores to increase even more, compared to similarity scores of Siamese pairs with already lower similarity scores. This is a self-reinforcing loop: in positive exemplar-target pairs, Siamese pairs with high similarity scores will get even higher similarity scores during training, therefore also higher attention weights and again a more negative gradient that will increase similarity scores for these locations even more. Similarly, similarity scores for Siamese pairs with already lower similarity score will be increased less compared to similarity scores of other Siamese pairs, causing lower attention weights and again a less negative gradient. See Figure \ref{fig:fig2} for a visualisation of the relation between the similarity scores and the gradient propagated to them. 

According to the above, it is enough to only make positive Siamese pairs have slightly larger similarity scores than negative Siamese pairs, and the self-reinforcing loop will increase this difference. The intuition here is that positive Siamese pairs have in general more in common with each other than negative pairs have in common with each other, in the sense that gradient updates based on minibatches will increase similarity scores of positive Siamese pairs more than negative ones. This difference between positive and negative Siamese pairs should be enough for the self-reinforcing loop to begin and eventually assign considerably larger similarity scores to positive Siamese pairs compared to negative ones. 

\subsection{One-Shot Learning} \label{sec:oneshot}
By conditioning the model only on the exemplar and the target example and not on any explicit class label, we allow the model to better generalise to unseen classes. Indeed, exemplars from unseen classes can share characteristics with already seen exemplars, aiding the model to make correct predictions for examples of unseen classes. This can be seen as the model implicitly learning to represent a class by its examples, then being able to generalise to similar classes in this class representation space. 

We define $T$ to be the uniform distribution of classes available at training and evaluation time. We denote with $X_L$ the distribution of exemplar-target pairs $(x, B)$ such that $x$ belongs to class $L$, and $B$ contains an instance of class $L$ with a probability of $0.5$. At training time, we sample a class from the distribution $T'$ that is uniform over the classes available at training time, and try to minimise the loss over exemplar-target pairs sampled from $X_L$ that corresponds to the appropriate class. Our goal of training is then choosing model parameters $\Theta$ such that 
\begin{equation} \label{eq:argmin}
\Theta = \arg \min \limits _\theta \mathbb{E}_{L \sim T'} [ \mathbb{E}_{(x, B) \sim X_L} \ell (x, B)]].
\end{equation}
Training the model with Eq. \ref{eq:argmin} should yield a model which performs well when sampling $L \sim T$, instead of $L \sim T'$, such that classes that were not seen in training time are included. We evaluate our model in the experiments section only on classes that were not included in the training set. For predicting well on unseen classes, our model does not need any fine-tuning that adjusts the model to the new classes.

\subsection{Detection} \label{sec:detections}
For a detection task, given a target example and a set of possible classes, we are interested in answering the question which of the possible classes appear in the target example, and what location they appear in. Our model takes as input a target example and an exemplar of one particular class, therefore in order to consider all possible classes we need to feed our trained model with the target example, together with an exemplar of each possible class. This can be done in a more computationally efficient way though, where $f_\theta (B_l)$ from Eq. \ref{eq:sim} is only computed once for each target location, and compared with the embeddings of all the different exemplars. 

For every exemplar-target pair $(x, B)$, we consider a possible detection with confidence $\hat{y}_{x, B}$ at the target location $l$ with the highest similarity score $s_l(x, B_l)$. Note that here we make a simplifying assumption that an instance of the class of the exemplar may appear in the target example at most once. This can be generalised to cases where this assumption is not true by emitting multiple detections if a high similarity score is present in multiple locations, but for simplicity we stick to the case where the assumption holds. See Section \ref{sec:eval} for more details about the process of emitting detections and their evaluation.

\section{Experiments} \label{sec:exp}
\subsection{Audio Data}
For performing weakly supervised one-shot detection in the audio domain, we tackle the task  of detecting audio keywords in longer audio utterances, also known as query-by-example spoken term detection \citep{DBLP:conf/asru/HazenSW09,DBLP:conf/asru/ParadaSR09,Woellmer09-RDK,Woellmer10-STD,DBLP:conf/icassp/ChenPS15}. In this task, every word represents a class. Each recording of an audio keyword is an exemplar of some class, and the longer audio utterances are the target examples. The goal is to determine whether the same word as the keyword appears in the utterance, and determine the bounding box location of the appearance, in seconds, in case the word appears in the utterance.

We construct a large-scale dataset for our keyword detection task, using two separate existing corpora: a speech recognition corpus and an audio keywords corpus. A total of 7258 recordings of audio keywords were downloaded from the Shtooka project website (\url{http://www.shtooka.net}), that result in 7258 different classes. The audio utterances we use are from the Librispeech corpus \citep{panayotov2015librispeech}. The Librispeech corpus contains 1000 hours of annotated English speech, from 2484 speakers. Each keyword and each utterance were allocated to either the training, the validation or the test set. 

Our training set is comprised of exemplar-target pairs (keyword-utterance pairs) where both the keyword and the utterance were allocated to the training set. The training set contains all keyword-utterance pairs where the keyword appears in the utterance, and we balance the training set with the same number of keyword-utterance pairs where the keyword does not appear in the utterance. In total, our training set contains 330,018 keyword-utterance pairs. 

Evaluation sets are constructed for an $N$-way one-shot detection. This means that in the test and validation sets for every target example $B$ there are $N$ exemplar-target pairs $\{(x_i, B)\} _{i=1} ^N\}$ such that $y_{x_i, B}=1$ for exactly one of the $N$ pairs. For constructing the evaluation sets (validation and test), we first add all keyword-utterance pairs that belong to the appropriate evaluation set, where the keyword appears in the utterance. For performing $N$-way one-shot detection, we use each utterance to create additional $N-1$ keyword-utterance pairs (with keyword again from the appropriate evaluation set) where the keyword does not appear in the utterance. Note that in all cases the keyword and the utterance are always from different recordings, made by different speakers. We represent the audio recordings as their spectrograms. Specifically, we apply we apply a short-time Fourier transform (STFT) over frames of 25\,ms, shifted by 10\,ms to every recording of a keyword or an utterance, to extract 201 magnitude features from every frame. A full description of the dataset creation process can be found in Section \ref{sec:ap_data} in the supplementary material.

\subsection{Computer Vision Data} \label{sec:image}
The Omniglot dataset \citep{lake2015human} contains images of 1623 different characters from 50 different alphabets. As discussed in the introduction, models that are not conditioned on class identities may demonstrate generalisation to classes unseen in training time, but this may require examples from a large number of classes to appear in training time. As the Omniglot dataset contains a large number of classes (characters), it is suitable for a one-shot learning and was used in previous works \citep{DBLP:conf/nips/VinyalsBLKW16,koch2015siamese}.

\begin{figure}
\centering
\begin{subfigure}{.33\textwidth}
  \centering
  \includegraphics{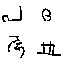}
  \caption{$n=2$}
\end{subfigure}
\begin{subfigure}{.33\textwidth}
  \centering
  \includegraphics{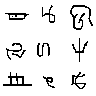}
  \caption{$n=3$}
\end{subfigure}
\begin{subfigure}{.33\textwidth}
  \centering
  \includegraphics{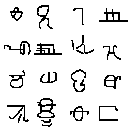}
  \caption{$n=4$}
\end{subfigure}
\caption{Sample tiled Omniglot images, used for the weakly supervised one-shot detection task}
\label{fig:omniglot}
\end{figure}

The dataset is comprised of images of single characters. To be suitable for a detection task, we use $n^2$ single character images to create one large square image that contains $n \times n$ images of different single characters. The $n^2$ images are non-overlapping and are arranged on a grid in a random order. Examples of such large images are found in Figure \ref{fig:omniglot}. All images of single characters are downsampled to $32 \times 32$ pixels to reduce the computational requirements of the model. The task is then, given an image of a single character (an exemplar) to determine whether a given large image (the target example) contains an instance of the same character as the exemplar, and localise this instance in the large image. Note that the exemplar's image never appears in the target example, but rather a different image, of the same character as the exemplar.

We allocate the alphabets of the Omniglot datasets into distinct training, validation and test splits. We use the same test alphabets as in the official split of this dataset. Each of the training, validation and test sets is comprised of exemplar-target pairs, such that all characters in both the exemplar and the target example are from the appropriate training, validation or test split. The training set for our task is comprised of exemplar-target pairs that are labelled either as positive, if the target example contains an instance of the same character as the exemplar, or negative, otherwise. For the evaluation sets (validation and test), for every image of a character, we use one exemplar-target pair such that an instance from the same character as the exemplar appears in the target image, and $N-1$ exemplar-target pairs in which the exemplar does not appear in the target image.

\subsection{Network Specifications}
Our model represents the exemplar and the location in the target example using a neural network $f_\theta$, as appears in Eq. \ref{eq:sim}. 
For both domains, we model $f_\theta$ as a CNN with the same structure, differing only by the convolution structure, one-dimensional convolutions for the audio data and two-dimensional convolutions for the image data. 
The network is similar to the networks used in \citet{keren2017tunable} and \citet{Keren17-POL} and has the following specifications. The network is comprised of eight convolutional layers. 
The network is comprised of eight convolutional layers, each using a kernel size of five ($5 \times 5$ for the image data), with a stride of one ($1 \times 1$ for the image data). The first four convolutional layers are comprised of 256 feature maps, while the last four convolutional layers are comprised of 512 feature maps. To reduce the representation size and enlarge the receptive field, after every second convolutional layer we apply a max-pooling operation with a kernel size and stride of two. Every convolutional layer is followed by a batch normalisation operation \citep{DBLP:conf/icml/IoffeS15} and rectified-linear activation function. 

For computing the cosine similarity between the exemplar and and the locations in the target example, we simply convolve the resulting exemplar representation with the representation of the target example, after normalising the $L^2$ norm of the exemplar representation and each location in the target example's representation. When computing the attention weights according to Eq. \ref{eq:softmax}, a softmax temperature of $T=\frac{1}{3}$ is used. The network is trained according to Eq. \ref{eq:argmin} using Stochastic Gradient Descent with a learning rate of 0.1 and minibatch size of 64 keyword-utterance pairs. All hyperparameters were tuned on a validation set. For the audio experiments, training is stopped when performance is best on the validation set. For the computer vision experiments, the model is first trained to determine the optimal number of training iterations, then retrained using the  training and validation sets to obtain the reported performance on the test set.

\begin{table*}[t]
\caption{Results for experiments in the audio and computer vision domains. Average precision (AP[\%]) and precision at given recall levels (Pr'@x[\%]) are reported on the different $N$-way test sets for the weakly supervised one-shot detection task, for our proposed attention similarity networks and the dynamic time warping and Exemplar-SVM baselines. Higher scores are better. For both domains, our attention similarity networks outperforms the baseline in all performance measures.}
\label{tab:results}
\vskip 0.15in
\begin{center}
\begin{small}
\begin{sc}
\begin{tabular}{llcccc}
\toprule
	Model & Set & AP & Pr'@0.5 & Pr'@0.9 & Pr'@0.99  \\ 
	\midrule[1pt]
    \multirow{3}{*}{Audio - attention similarity networks}
     & $10$-way & 42.6 & 73.1 & 25.9 & 12.7  \\ 
     & $20$-way & 38.3 & 50.9 & 16.0 & 6.5  \\ 
     & $50$-way & 23.6 & 29.9 & 5.8 & 2.6  \\    
    \cdashline{1-6}
    \rule{0pt}{2.5ex}
    \multirow{3}{*}{Audio - Dynamic Time Warping}      
     & $10$-way & 8.9 & 14.6 & 11.1 & 10.4  \\
     & $20$-way & 6.0 & 7.8 & 5.8 & 5.5  \\
     & $50$-way & 3.7 & 3.3 & 2.4 & 2.3  \\ 
     \midrule[1pt]
    \multirow{3}{*}{Image - attention similarity networks}      
     & $5$-way  & 58.0 & 73.1 & 39.4 & 24.5  \\
     & $10$-way & 49.3 & 63.1 & 22.0 & 12.0  \\
     & $20$-way & 37.2 & 41.5 & 11.9 & 6.2  \\ 
    \cdashline{1-6}
    \rule{0pt}{2.5ex}
    \multirow{3}{*}{Image - Exemplar-SVM}      
     & $5$-way  & 31.1 & 40.3 & 23.4 & 20.5  \\
     & $10$-way & 24.8 & 23.0 & 12.0 & 10.3  \\
     & $20$-way & 19.6 & 12.4 & 6.1 & 5.2  \\ 
     \midrule[1pt]
\end{tabular}
\end{sc}
\end{small}
\end{center}
\vskip -0.1in
\end{table*}

\subsection{Evaluation} \label{sec:eval}
We evaluate our weakly supervised one-shot detection model on detection tasks from the audio and computer vision domains. Note that even though in the case of the datasets we use in this work we can acquire bounding box labels, such labels are in many scenarios not available for the majority of the data (for example, for detection in real-world images), making this showcase for weakly supervised learning important. Bounding box labels are not used in our experiments for training, but rather only for evaluation of the trained model. We leave weakly supervised one-shot detection with real-world image data for future work. 

For the audio experiments, we use dynamic time warping (DTW) as a baseline to compare our proposed method to. DTW matches sequences of different lengths with each other \citep{Woellmer09-AMD}. As the duration of the articulation of a word usually differs between speakers and situations, DTW is a well-established approach for query-by-example spoken term detection \citep{Joder12-TTC}. As it does not require any training phase, it is suitable for our one-shot query-by-example paradigm. For the computer vision experiments, we use Exemplar-SVM as a baseline. In this method, an SVM with a linear kernel is trained for each test exemplar,  separating this exemplar from all training set exemplars. This trained model is then convolved with the target example for emitting similarity scores for the different locations. We did not find in the literature any neural network-based methods that are suitable for weakly supervised one-shot detection and can be directly compared with our method. For a more detailed description of the two baseline methods, see the Sections \ref{sec:ap_dtw} and \ref{sec:ap_SVM} in the supplementary material. 


All methods we consider output for an exemplar-target pair $(x,B)$ a possible detection with confidence $\hat{y}_{x, B}$ at location $l$. We emit a detection for $(x,B)$ if $\hat{y}_{x, B} > t$, where $t$ is a threshold chosen according to detection performance on the validation set, for each method separately. For the audio experiments, we shift all start and end points of all detections' locations by constants $a$ and $b$, that are again chosen according to detection performance on the validation set, for each method separately. For the image data experiments, we do not consider this shift postprocessing as single character images can appear only in a small number of locations in the larger image. 

Given a method's detections and the ground truth bounding boxes, we compute our main performance measure in terms of average precision, as done for object detection models in computer vision, using an intersection over union (IoU) threshold of 0.5 \citep{DBLP:journals/ijcv/EveringhamGWWZ10}. The only difference between the performance measure in \citet{DBLP:journals/ijcv/EveringhamGWWZ10} and our performance measure is that in our case we compute the average precision (AP) for detection of all classes together, instead of averaging the AP computed for each class separately (mAP). The reason for this deviation from the well established performance measure is that in general we have a large number of classes but a small number of ground truth bounding boxes per class.

In addition to the detection performance evaluated by AP, as every method is trained to perform binary classification of exemplar-target pairs $(x, B)$ according to the existence of the exemplar $x$ in the target example $B$, we additionally evaluate all models' binary classification test performance by computing precision at recall levels of 0.5, 0.9 and 0.99, where the $\hat{y}_{x,B}$ is the score assigned to each keyword-utterance pair.


For the experiments with computer vision data, we construct the larger images with $n \in \{2,3,4\}$, as explained in Section \ref{sec:image}. In addition, we construct three different test sets, for $5$-way, $10$-way and $20$-way weakly supervised one-shot detection. For the experiments with audio data, we use three different test sets for, for $10$-way, $20$-way and $50$-way weakly supervised one-shot detection. Results for audio experiments and the computer vision experiments with $n=3$ are reported in Table \ref{tab:results}. Additional results for the computer vision experiments with $n=2$ and $n=4$ are of the exact same quality and are provided in Section \ref{sec:ap_additional} in the supplementary material. Our proposed attention similarity networks considerably outperformed the baseline methods for both the audio and computer vision domains, for all test sets. Specifically, in the audio domain our proposed model yielded AP scores of 42.6\%, 38.3\% and 23.6\% for the detection task over 10, 20 and 50 unseen classes, compared to AP scores of 8.9\%, 6.0\% and 3.7\% with DTW. In the computer vision domain, our proposed method yielded AP scores of 58.0\%, 49.3\% and 37.2\% for the detection task over 5, 10 and 20 unseen classes, compared to AP scores of 31.1\%, 24.8\% and 19.6\% with the Exemplar-SVM.

Our proposed attention similarity networks managed to simultaneously identify and localise instances of classes unseen at training time. AP results show that it is indeed feasible to generalise to unseen classes when using a model that is not conditioned on class identities, and that localisation information can be learnt in a weakly supervised manner, when pairing a Siamese similarity network with an attention mechanism.

\begin{figure}[ht]
\vspace{-0.3cm}
\begin{center}
\centerline{\includegraphics[scale=0.2]{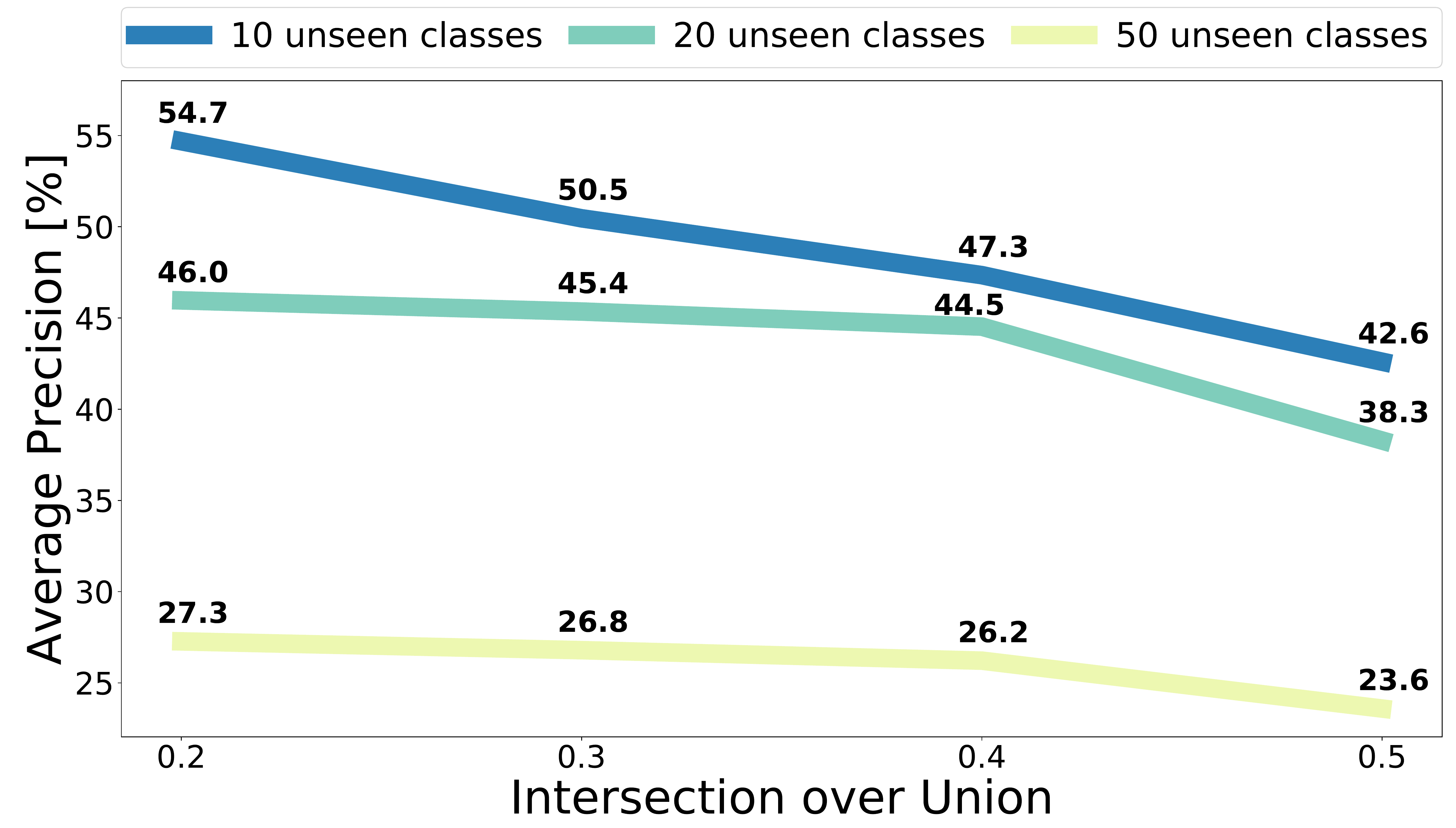}}
\caption{Average precision (AP) of the attention similarity networks for the audio keywords detection task for the $10$-way, $20$-way and $50$-way weakly supervised one-shot detection with different intersection over union (IoU) thresholds.}
\label{fig:fig3}
\end{center}
\vskip -0.2in
\end{figure}

To further investigate possible future improvements for our architecture, we compute AP for our trained attention similarity networks using different IoU thresholds. Results are depicted in Figure \ref{fig:fig3}. For the audio data and $10$-way one-shot detection, we found that reducing the IoU threshold from 0.5 to 0.4 improved the AP from 42.6 to 47.3. Further reducing the IoU to 0.3 and 0.2 resulted in AP values of 50.5 and 54.7 respectively. Similar findings were found for the $20$-way and $50$-way one-shot detection tasks. These results motivate future work and improvements for this model, as they show that further refining the detection locations is expected to improve the detection performance for unseen classes. 


\section{Conclusion} \label{sec:conclusion}
We further explored and established generalisation to unseen classes in models that are not conditioned on class identities, that may potentially be a key paradigm in machine learning for both the one-shot and the non-one-shot learning settings. 
For the task of weakly supervised one-shot detection, we presented the attention similarity networks model, that manages to simultaneously identify and localise an instance on a class unseen in training time. We experimented with datasets from the audio and computer vision domains, and found our model to considerably outperform the baseline methods. We conclude that models with no conditioning on class identities are appropriate for performing one-shot detection tasks. Furthermore, we conclude that pairing a Siamese similarity network with an attention mechanism enables learning of similarity between object parts in an weakly supervised manner. 

Future work should further explore the advantages of models with no conditioning on class identities for one-shot tasks in addition to non-one-shot tasks, as such models can facilitate knowledge transfer between classes in both cases. In addition, weakly supervised one-shot detection models should be further developed, refining their outputs using model confidence predictions \cite{keren2018calibrated} and incorporating adjusted variants of architectural developments from non-one-shot and fully supervised computer vision object detection such as \citet{DBLP:conf/nips/RenHGS15}.

\bibliography{oneshot}
\bibliographystyle{icml2018}

\clearpage
\begin{center}
\hrule

\vspace{0.5em}
{\large{Supplementary material}}
\vspace{0.5em}
\hrule
\end{center}

\appendix

\section{Audio data} \label{sec:ap_data}
We include a full description of the audio data creation. 
We construct a large-scale dataset for our keyword detection task, using two separate existing corpora: a speech recognition corpus and an audio keywords corpus. The audio keywords were downloaded from the Shtooka project website (\url{http://www.shtooka.net}). All keywords are in English, and are less than one second long. Every audio keyword is unique, in the sense that no word appears in two different audio recordings. Each keyword was allocated to either the training, the validation or the test set. 
By doing so, we make sure we evaluate the model on detection of keywords that were not seen during the training phase, which is a one-shot learning task. The textual from of words can appear as a part of other, longer words (for example, `the' is a part of `their' and `further'), which results in an undefined desired behaviour for the model. For this reason, we chose not to use short words, that are more prone to this issue, and we only use words that consist of four letters or more. In total, our training set contains 5442 keywords (classes), and the validation and the test set contain 908 keywords each.

All audio utterances we use are from the Librispeech corpus \citep{panayotov2015librispeech}. The Librispeech corpus contains 1000 hours of annotated English speech, from 2484 speakers. Each utterance in the Librispeech corpus was allocated to one of our training, validation or test sets, according to the official split of this corpus, which is gender balanced. We cut each utterance to be exactly five seconds long.

Our training dataset is comprised of keyword-utterance pairs, with binary labels that indicate the existence of the keyword in the utterance. We use the term \emph{positive pairs} to refer to keyword-utterance pairs where the keyword appears in the utterance (according to the utterances' transcriptions), and \emph{negative pairs} to refer to other pairs. Note that the keyword and the utterance are always from different recordings, made by different speakers. For constructing our training set, we use all positive keyword-utterance pairs where both the keyword and utterance were allocated to the training set. In order to balance the two classes in the training set, we make sure every keyword appears in the same number of positive and negative pairs. We do this by randomly sampling a number of training utterances that do not contain the keyword and add the resulting negative pairs to the training set. In total, our training set contains 330,018 keyword-utterance pairs. 

For constructing the validation and test sets, we first add all positive keyword-utterance pairs that belong to the appropriate evaluation set (validation or test). For evaluating the model on detection over a number of unseen possible classes (different unseen classes every time though), for every positive keyword-utterance pair in the evaluation set we add another $n$ negative keyword-utterance pairs that comprise of the same utterance as in the positive pair. The keywords and utterances in the negative pairs we add are also from the same set (validation or test) as the ones in the positive pair. Note that the more negative keyword-utterance pairs we add to the model, the more false positives we are likely to find, that should impair the overall performance of the detection model. We create the validation and test set as described above with $n \in \{10, 20, 50\}$, and we name the resulting test sets $test10$, $test20$ and $test50$ respectively. See Section \ref{sec:eval} for details about the detection task evaluation method.

Labels regarding the temporal location of a keywords in an utterance were extracted using forced alignment. Specifically, we used the Montreal Forced Aligner \citep{mcauliffe2017montreal} with default parameters. These labels were used for creating ground truth bounding boxes, that are used when evaluating the detection model on the validation and test sets. However, these labels were not used at training time, as in this work we consider a weakly supervised model. 

As some words are more common than others, every keyword has a different number of keyword-utterance pairs that it is a part of, which can result in a small number of keywords dominate the training or evaluation procedure.  To counter this effect, in each of the training, validation and test sets, we count the number of keyword-utterance pairs that each keyword appears in, and we use only keywords that are below the 85th percentile in this count. Overall, our training, validation and test sets contain 3592, 259 and 285 audio keywords respectively.

\section{Dynamic Time Warping} \label{sec:ap_dtw}
We use dynamic time warping (DTW) as a baseline to compare our proposed method to in our experiments with audio data. DTW matches sequences of different lengths with each other \citep{Woellmer09-AMD}. As the duration of the articulation of a word usually differs between speakers and situations, DTW is a well-established approach to query-by-example spoken term detection \citep{Joder12-TTC}. As it does not require any training phase, it is suitable for our one-shot query-by-example paradigm. 
As previously done for spoken term detection with DTW \citep{Joder12-TTC}, we use Mel-frequency cepstral coefficients (MFCCs) as acoustic features. We use the coefficients one to twelve, extracted from frames of 20\,ms shifted by 10\,ms with the toolkit \textsc{openSMILE} \citep{Eyben13-RDI}. 
The DTW algorithm finds the shortest path between the MFCC representation over time of each keyword and a given segment from an utterance. DTW then returns a cost value describing the sum of the Euclidean distances of the shortest path between the sequences. This cost function $C$ is converted to the probability of a match, which we use as the similarity measure between the keyword and the location in the utterance
\begin{equation}
s_l = \exp{\frac{-C}{\sigma}},
\end{equation}
with the parameter $\sigma$, where $50$ was found to yield best results on the validation set. The confidence was computed for each possible location within the utterance, \ie for each 10\,ms step as this is the hopsize of the MFCC feature vectors. For emitting detections, for an exemplar-target pair $(x,B)$ we define $\hat{y}_{x, B} = \max \limits _l s_l(x, B)$, and consider this a possible detection at location $l$ where the maximum was acquired with confidence $\hat{y}_{x, B}$.

\section{Exemplar-SVM} \label{sec:ap_SVM}
As a baseline for the computer vision experiments, we used an Exemplar Support Vector Machine (Exemplar-SVM) based on Histogram of Oriented Gradients (HOG)-features.
The Exemplar-SVM approach for the purpose of object recognition in images has been proposedin \citet{Malisiewicz11-EOE}. In \citet{Malisiewicz11-EOE}, the authors use the HOG-representation of objects in images and trained several SVM classifiers, with only a single positive instance (object is present in a bounding box) and a large number of negative ones (object is not present) for each SVM, and combined the outputs of the models within an ensemble. HOG features were also used for the recognition of handwritten characters in \citet{Surinta15-RHC}. 

As opposed to the task from \citet{Malisiewicz11-EOE}, in this work we take into account only a single positive exemplar. We tune the complexity hyperparameters for the exemplar and the negative sample on the validation set to choose the values of $10$ and $1e^{-4}$ respectively. As can be expected, complexity for the exemplar class is much larger than that of the negative sample. For each single character image from the test set characters, we train an SVM with a linear kernel to separate this image from the entire training set exemplars (the training set contains other characters then the test set). The SVM is trained on HOG features extracted from each image, as those were found to yield better results then raw pixels representation in initial experiments. For every exemplar-target pair $(x, B)$, we convolve the trained SVM for the single character image $x$ with the larger image $B$, to get the similarity scores $s_l(x, B)$ for the different locations. For emitting detections, for an exemplar-target pair $(x,B)$ we define $\hat{y}_{x, B} = \max \limits _l s_l(x, B)$, and consider a possible detection at location $l$ where the maximum was acquired with confidence $\hat{y}_{x, B}$.

For the SVM training we use the toolkit \textsc{Liblinear}~\citep{Fan08-LAL} employing the dual L2-regularised logistic regression solver, a bias of $1$ and the weighted complexities as described above. To extract the HOG features, the \textsc{scikit-image} library~\citep{Van14-SIP} is used with $4 \times 4$ pixels per cell.

\section{Additional Results for the Computer Vision Experiments} \label{sec:ap_additional}
We include additional results for the computer vision experiments, when constructing the larger Omniglot images with $n=2$ and $n=4$, as described in Section \ref{sec:image}. The results table compare our proposed attention similarity networks and the Exemplar-SVM baseline. Results for $n=2$ and $n=$ are found in Table \ref{tab:resn24}. Results are of the same nature as in Table \ref{tab:results}. See Section \ref{sec:eval} for the results analysis. 

\begin{table*}[t]
\caption{Results for experiments in the audio and computer vision domains ($n=2$ and $n=4$). Average precision (AP[\%]) and precision at given recall levels (Pr'@x[\%]) are reported on the different $N$-way test sets for the weakly supervised one-shot detection task, for our proposed attention similarity networks and the dynamic time warping and Exemplar-SVM baselines. Higher scores are better. For both domain, our attention similarity networks outperforms the baseline in all performance measures.}
\label{tab:resn24}
\vskip 0.15in
\begin{center}
\begin{small}
\begin{sc}
\begin{tabular}{llcccc}
\toprule
	Model & Set & AP & Pr'@0.5 & Pr'@0.9 & Pr'@0.99  \\ 
	\midrule[1pt]
	$n=2:$ \\
    \multirow{3}{*}{Image - attention similarity networks}      
     & $5$-way  & 75.4 & 89.3 & 52.3 & 27.0 \\
     & $10$-way & 67.1 & 79.2 & 36.3 & 14.9 \\
     & $20$-way & 54.4 & 61.8 & 18.1 & 7.4 \\
    \cdashline{1-6}
    \rule{0pt}{2.5ex}
    \multirow{3}{*}{Image - Exemplar-SVM}      
     & $5$-way  & 42.6 & 50.8 & 25.0 & 20.7 \\
     & $10$-way & 33.8 & 31.2 & 12.9 & 10.4 \\
     & $20$-way & 26.6 & 17.7 & 6.6 & 5.2 \\
     \midrule[1pt]

     $n=4:$ \\
    \multirow{3}{*}{Image - attention similarity networks}      
     & $5$-way  & 43.2 & 56.7 & 27.6 & 22.3 \\
     & $10$-way & 32.1 & 34.4 & 13.7 & 10.9 \\
     & $20$-way & 23.8 & 19.8 & 7.2 & 5.2 \\
    \cdashline{1-6}
    \rule{0pt}{2.5ex}
    \multirow{3}{*}{Image - Exemplar-SVM}      
     & $5$-way  & 26.3 & 34.8 & 22.6 & 20.4 \\
     & $10$-way & 20.8 & 19.1 & 11.5 & 10.2 \\
     & $20$-way & 16.4 & 10.0 & 5.8 & 5.1 \\
     \midrule[1pt]
\end{tabular}
\end{sc}
\end{small}
\end{center}
\vskip -0.1in
\end{table*}

\end{document}